\documentclass[conference]{IEEEtran}
\IEEEoverridecommandlockouts

\usepackage{cite}
\usepackage{amsmath,amssymb,amsfonts}
\usepackage{graphicx}
\usepackage{booktabs}
\usepackage{xcolor}
\usepackage{url}
\usepackage[hidelinks]{hyperref}
\def\BibTeX{{\rm B\kern-.05em{\sc i\kern-.025em b}\kern-.08em T\kern-.1667em\lower.7ex\hbox{E}\kern-.125emX}}

\begin{document}

\title{AgriDefend: A \$205 Offline Edge Device that Detects\\
Crop Stress Before Visible Symptoms via Multimodal AI Fusion}

\author{%
\IEEEauthorblockN{George Xu}
\IEEEauthorblockA{\textit{Independent Researcher}\\
Bothell, WA, USA\\
georgexu5588@gmail.com}
\and
\IEEEauthorblockN{Nikhil Vincent}
\IEEEauthorblockA{\textit{Independent Researcher}\\
Bothell, WA, USA\\
nikhil.vincent.v@gmail.com}
\and
\IEEEauthorblockN{Elizabeth S. Forbes}
\IEEEauthorblockA{\textit{Univ.\ of Michigan, School for}\\
\textit{Environment and Sustainability}\\
Ann Arbor, MI, USA\\
forbeselizabeth.s@gmail.com}
}

\maketitle

\begin{abstract}
Most in-field crop-monitoring devices sense a single modality---either a
camera looking at the canopy or a probe reading the soil---and so cannot
detect stress that begins below ground before it becomes visible above it.
We present AgriDefend, a self-contained consumer-grade edge device that
fuses both. A MobileNetV2 convolutional neural network (CNN) analyses canopy
imagery, and a Random Forest classifier integrates trend features from a
single soil probe: moisture, the rate of change in pH and in N, P, K, and
belowground CO$_2$ relative to ambient (a proxy for soil respiration). A
logistic-regression meta-learner fuses the two into a five-class
crop-health prediction, all on a Raspberry\,Pi\,5 with no cloud or cellular
link. On a controlled greenhouse drought trial with wild radish, the fused
model reached 87\% accuracy ($R^2=0.88$, MAE $=0.05$) at 1.4\,s INT8
TensorFlow Lite inference, and its soil signals preceded visible wilting by
about four days. The complete bill of materials is \$205, roughly a fifth of
comparable commercial nodes, positioning multimodal, sustainable crop
sensing as a practical consumer-electronics product for
connectivity-limited settings.
\end{abstract}


\section{Introduction}
Agriculture is the largest human use of freshwater---about 70\% of global
withdrawals---and a major consumer of synthetic nitrogen and phosphorus
fertilizers, while extreme weather and soil degradation add to the stress
crops experience in the field \cite{ref11,ref6}. Precision-agriculture
sensors help growers respond, but two barriers keep them out of many hands.
First, most commercial systems assume the cloud: they need continuous
cellular or broadband connectivity to push data to a server, which rural
areas often lack. Second, commercial nodes frequently cost more than
\$1{,}000 each.

There is also a diagnostic gap. Vision models detect visible leaf disease
and water stress from RGB imagery \cite{ref14,ref9}, and tabular models map
soil chemistry to nutrient status \cite{ref10}, but each uses a single
modality and cannot relate what happens in the soil to what the canopy shows.
Because visible stress lags the deficit that causes it, one modality gives
little warning. Multimodal late fusion---running a separate model per
modality and combining their outputs \cite{ref2}---is well suited to an
image stream and a low-rate soil stream that are sampled asynchronously.

AgriDefend brings these ideas together in a low-cost consumer device. Because
single-modality vision already classifies \emph{visible} stress at 85--99\%
\cite{ref14,ref9}, our contribution is not higher accuracy on an
already-symptomatic plant but two things a lone camera or probe cannot give:
(i) \emph{lead time}---an asynchronous late-fusion model in which the soil
branch surfaces a deficit before the canopy shows it---and (ii)
\emph{stress-type discrimination} across five states (Healthy, Water Stress,
Nutrient Deficient, pH Imbalance, Disease), delivered on (iii) an offline-first
hardware platform that runs the model for \$205 with no cloud or cellular
dependency. To our knowledge this combination---canopy--soil fusion,
pre-symptomatic detection, and fully offline low-cost edge deployment in one
device---has not been shown before. By \emph{offline-first} we mean a device
that senses, infers, and serves results locally and treats any network
connection as optional.

\section{Related Work}
Machine learning for crop health has largely grown along two separate tracks.
On the vision side, convolutional networks classify visible disease and water
stress from RGB imagery at 85--99\% accuracy under controlled conditions
\cite{ref14,ref9}, and lightweight backbones such as MobileNet have made
on-device inference practical. On the sensing side, tabular models map soil
moisture and NPK readings to nutrient status or yield \cite{ref10}. Both lines
are mature, yet each sees only half of the plant--soil system: a camera cannot
see a developing root-zone deficit, and a soil probe cannot see a leaf lesion.
Multimodal learning is the natural bridge \cite{ref2}, and among the usual
early-, late-, and hybrid-fusion strategies, late fusion fits best when the two
streams are sampled at very different rates and must be combined on
resource-constrained hardware. Existing multimodal agriculture systems,
however, tend to assume cloud connectivity for the fusion step or rely on
research-grade instrumentation rather than a self-contained, low-cost field
device. AgriDefend targets exactly that gap: canopy--soil late fusion,
pre-symptomatic detection, and fully offline inference in one \$205 unit.

\section{System Design}

\subsection{Hardware}
\begin{figure}[t]
  \centering
  \includegraphics[width=0.85\linewidth]{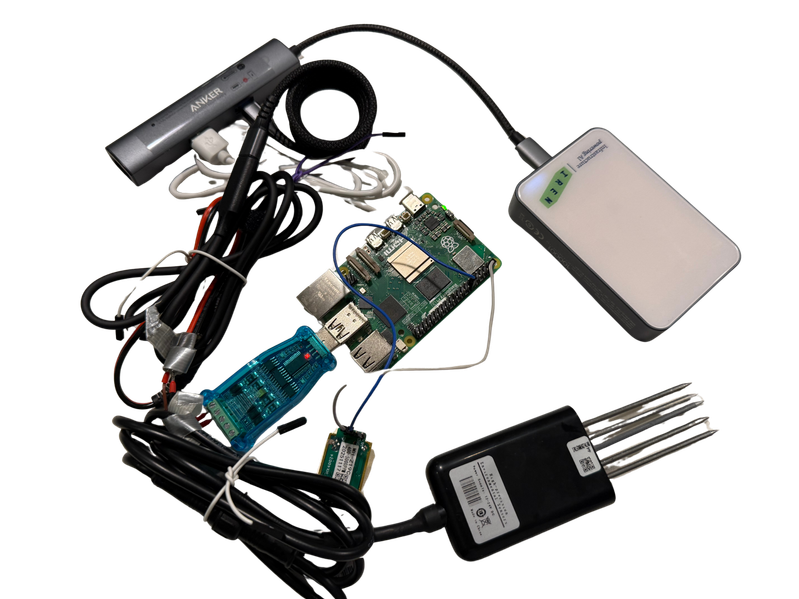}
  \caption{The AgriDefend node: Arducam camera (top left),
  Raspberry\,Pi\,5, the ADS1115/MAX485 bridge board (blue), an RS-485
  7-in-1 soil probe (right), and a battery pack. The bridge lets the Pi
  read the Modbus soil sensors over RS-485.}
  \label{fig:bridge}
\end{figure}

A waterproof enclosure houses an RS-485 soil probe, an RS-485 Modbus RTU
sensor, and the cabling. The Raspberry\,Pi\,5 cannot read analog signals or
speak RS-485 over its GPIO, so a bridge (Fig.~\ref{fig:bridge}) uses an
ADS1115 for analog-to-digital conversion over I\textsuperscript{2}C and a
MAX485 to convert TTL-level UART to RS-485. An Arducam captures the visual
stream. The Pi also acts as its own Wi-Fi access point (\texttt{hostapd},
\texttt{dnsmasq}); a React Native app reads live data from a local Flask API
when a user joins the device's network---no cell service or cloud account
needed.

\begin{table}[t]
  \centering
  \caption{Bill of Materials (per node).}
  \label{tab:bom}
  \begin{tabular}{lrr}
    \toprule
    \textbf{Part} & \textbf{Qty} & \textbf{Cost} \\
    \midrule
    Raspberry Pi 5 & 1 & \$60 \\
    Arducam module & 1 & \$25 \\
    RS-485 7-in-1 soil sensor & 1 & \$75 \\
    MH-Z19C CO$_2$ sensor & 1 & \$20 \\
    ADS1115 \& MAX485 ICs & 1 set & \$10 \\
    3D-printed case & 1 & \$15 \\
    \midrule
    \textbf{Total} & & \textbf{\$205} \\
    \bottomrule
  \end{tabular}
\end{table}

\subsection{Multimodal Late-Fusion Model}
The visual branch captures RGB frames at 1\,FPS, pre-processes with OpenCV
\cite{ref3} (HSV conversion, gamma correction $\gamma=1.2$), and feeds a
MobileNetV2 backbone with a five-class head, producing a probability vector
$\mathbf{V}\in\mathbb{R}^5$ over \{Healthy, Water Stress, Nutrient Deficient,
pH Imbalance, Disease\}. The soil branch samples RS-485 telemetry every
5\,min into
$\mathbf{S}=[\dot{p}\mathrm{H},\Delta\mathrm{CO_2},\dot N,\dot P,\dot K,\theta]$,
where dots denote temporal slopes and $\theta$ is volumetric moisture. Because
low-cost sensors are noisy (NPK error up to $\pm15\%$), the branch uses trend
features: OLS slopes over sliding windows, and a differential CO$_2$ signal
\begin{equation}
  \Delta\mathrm{CO_2}(t)=\mathrm{CO_{2,sub}}(t)-\overline{\mathrm{CO_{2,amb}}}(t)
\end{equation}
that isolates root-zone activity from atmospheric swings. A Random Forest
($n_\text{est}=200$, depth 15, \texttt{class\_weight=balanced}) \cite{ref4}
maps $\mathbf{S}$ to $\mathbf{R}\in\mathbb{R}^5$. A logistic-regression
meta-learner fuses $\mathbf{V}$ and $\mathbf{R}$; we keep this combiner
deliberately lightweight rather than using an attention- or transformer-based
fusion head, because the branches sample at very different rates and are fused
asynchronously, a linear combiner cannot overfit a small field dataset the way a
high-capacity head can, and it adds negligible latency while yielding
interpretable per-modality weights. The most recent soil vector is carried
forward and fused with each incoming frame, so a prediction is available at
camera rate. The final class is $\hat y=\arg\max\mathbf{F}$. To rule out
train-on-test leakage, the meta-learner is fit only on out-of-fold predictions
from the training partition, generated by 5-fold cross-validation within the
training set; the validation and test partitions are never seen during fusion
training. Fusion uses cross-entropy loss with balanced class weighting.

\subsection{Offline operation and interface}
Everything a grower needs runs on the node itself. The Raspberry\,Pi\,5 hosts
its own Wi-Fi access point (\texttt{hostapd}, \texttt{dnsmasq}), so a phone
joins the device directly with no router, SIM card, or cloud account; a React
Native app then reads live predictions and soil-sensor trends from a local
Flask API. Each node is fully self-contained, and a field of nodes has no
shared gateway or server to fail. Beyond the predicted class, the app surfaces
\emph{which} modality drove an alert---a direct benefit of the interpretable
per-modality fusion weights---so a grower can tell whether a warning came from
the canopy or from the soil before acting on it. This offline-first,
human-in-the-loop design is what makes AgriDefend usable as a consumer product
in connectivity-limited settings rather than a lab demonstrator.

\section{Evaluation}

\subsection{Accuracy}
We trained on $n=3{,}165$ wild-radish (\textit{Raphanus raphanistrum})
images (80/20 split, augmentation on training only) and evaluated in a
controlled greenhouse drought trial (one node per pot across 10 pots, over
7 days after final watering). Accuracy is measured against expert visual
scoring together with gravimetric soil-moisture labels. The fused model
reached 87\% overall accuracy ($R^2=0.88$, MAE $=0.05$), with 92\% precision
on the Healthy class. An ablation confirms fusion helps: CNN-only 78\%,
RF-only 75\%, full fusion 87\%. The nine-point gain of fusion over the stronger
single branch is larger than the gap between the two branches themselves, which
suggests the image and soil streams carry complementary rather than redundant
information---the meta-learner recovers cases that neither modality classifies
confidently alone. The 92\% precision on the Healthy class also matters in
practice: a field-warning tool that cries wolf quickly loses a grower's trust,
and the system stays conservative about declaring a plant stressed.

\begin{figure}[t]
  \centering
  \includegraphics[width=0.9\linewidth]{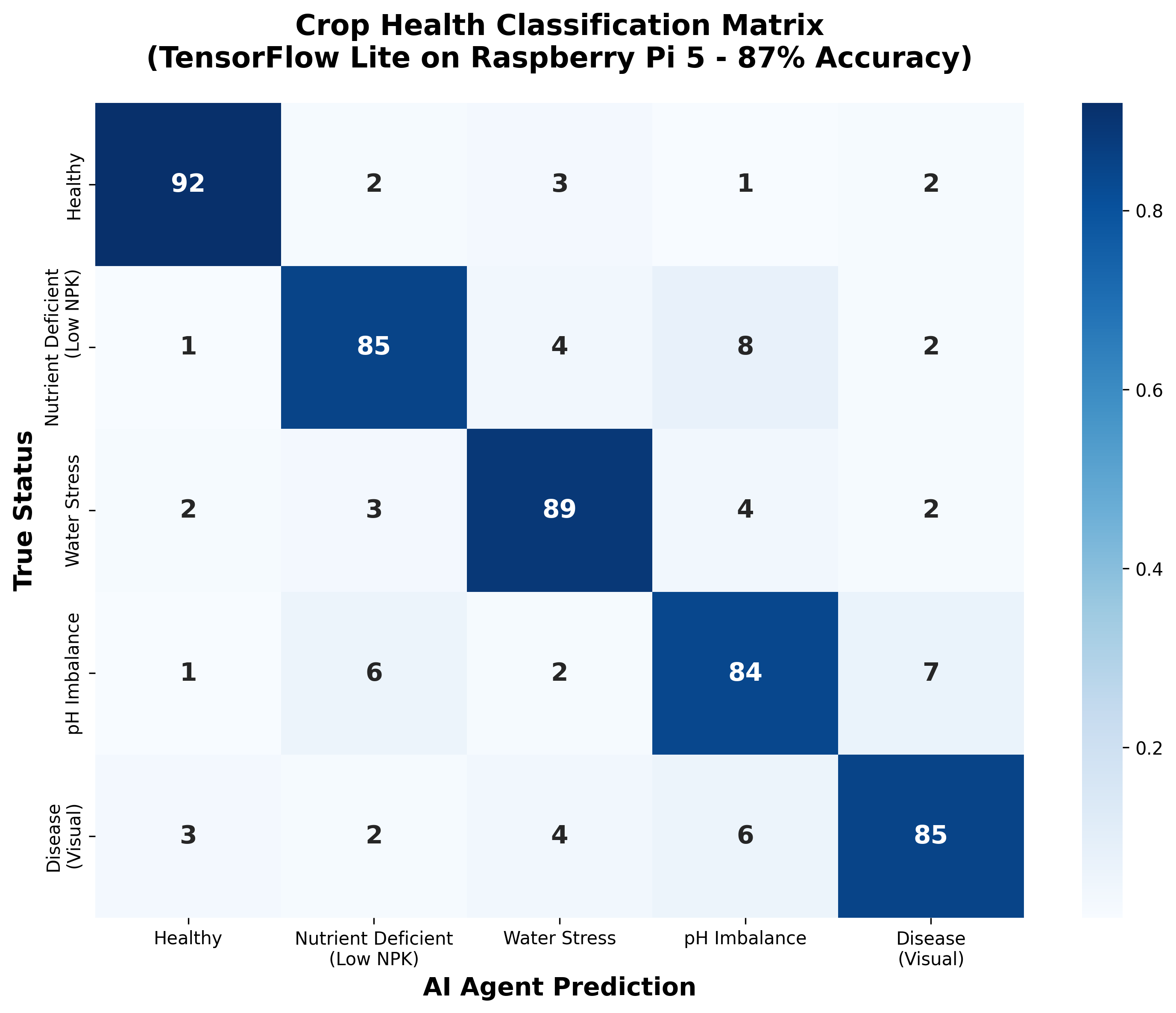}
  \caption{Five-class confusion matrix, TensorFlow Lite on Raspberry\,Pi\,5
  (87\% overall accuracy).}
  \label{fig:cm}
\end{figure}

\subsection{Edge Latency}
On an identical input ($224\times224$ RGB frame plus 6-feature soil vector),
averaged over 50 runs, inference latency was 3.5\,s via a cloud endpoint,
2.8\,s for full-precision TensorFlow on a Raspberry\,Pi\,4, and 1.4\,s
($\pm0.08$\,s) for INT8 TensorFlow Lite on a Raspberry\,Pi\,5. Only the last
met our 2.0\,s target for real-time field use \cite{ref16}.

\subsection{Early Warning}
\begin{figure}[t]
  \centering
  \includegraphics[width=0.9\linewidth]{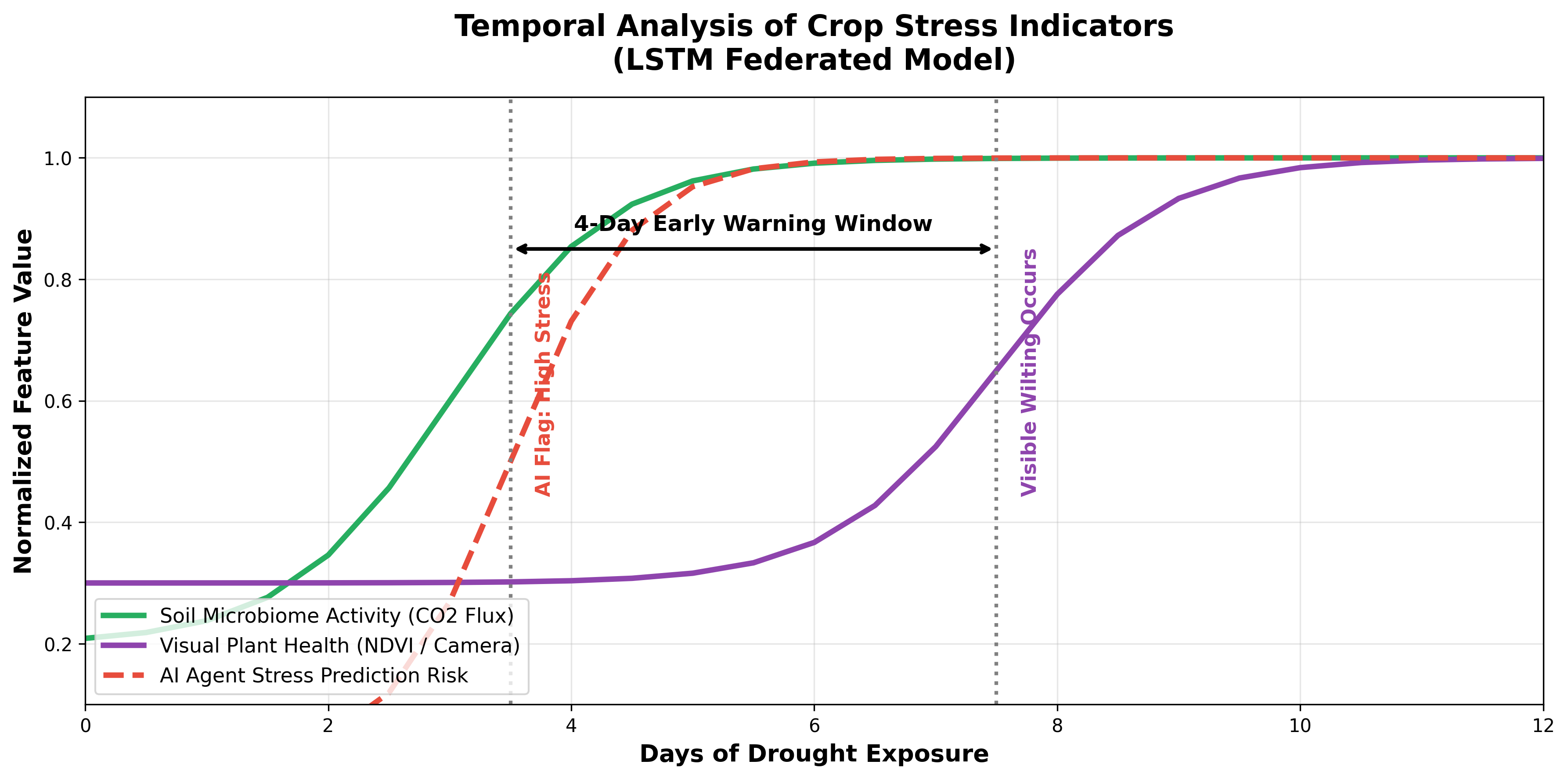}
  \caption{Soil CO$_2$ flux and moisture deficit rise several days before
  visible wilting.}
  \label{fig:timeline}
\end{figure}

In the drought trial, the soil branch flagged rising differential CO$_2$ and
an accelerating moisture deficit at $\sim$day~3.5, while the CNN branch did
not register visible wilting until $\sim$day~7.5 (Fig.~\ref{fig:timeline})---
an early-warning window of about four days. This ordering is what root-zone
physiology predicts: as the rhizosphere dries, falling root hydraulic
conductance triggers stomatal closure before turgor loss becomes visible, so a
soil-side moisture and respiration signal necessarily leads the canopy symptom
\cite{refphys1,refphys2}.

\section{Discussion}
A single \$205 node costs about a fifth of a commercial competitor, so a
grower could deploy a five-node mesh for less than one commercial unit and
map where stress is developing across an acre. Because each unit is
decentralized and offline-capable, an array of nodes has no single gateway
to fail. The contribution is not to beat single-modality accuracy on already-visible
stress---a well-solved problem \cite{ref14,ref9}---but to add the lead time and
stress-type resolution a single modality cannot.

\subsection{Limitations and future work}
We are careful not to over-claim from these numbers. Validation is
greenhouse-based and covers a single species (wild radish) in one drought
trial across ten pots, with ground truth from expert visual scoring and
gravimetric soil-moisture labels rather than instrumented plant-physiology
measurements. The five-class scheme is also dominated by water stress in this
trial; the remaining classes are represented but not yet stress-tested across
many field conditions. The next steps, which we are actively pursuing, are
open-field deployment at multiple sites, additional crop species,
per-stress-type lead-time characterization, and quantitative physiological
ground truth such as leaf water potential and tissue nutrient assays. On the
hardware side, LoRaWAN meshing would free nodes from local Wi-Fi range
\cite{ref15} and enable the distributed, gateway-free arrays that motivate the
offline-first design in the first place.



\begin{thebibliography}{00}
\bibitem{ref2} T.~Baltru\v{s}aitis, C.~Ahuja, and L.-P.~Morency, ``Multimodal machine learning: A survey and taxonomy,'' \textit{IEEE Trans. Pattern Anal. Mach. Intell.}, vol.~41, no.~2, pp.~423--443, 2018.
\bibitem{ref3} G.~Bradski, ``The OpenCV library,'' \textit{Dr. Dobb's Journal of Software Tools}, 2000.
\bibitem{ref4} L.~Breiman, ``Random forests,'' \textit{Machine Learning}, vol.~45, no.~1, pp.~5--32, 2001.
\bibitem{ref6} J.~A.~Foley et al., ``Solutions for a cultivated planet,'' \textit{Nature}, vol.~478, pp.~337--342, 2011.
\bibitem{ref9} A.~Kamilaris and F.~X.~Prenafeta-Bold\'u, ``Deep learning in agriculture: A survey,'' \textit{Comput. Electron. Agric.}, vol.~147, pp.~70--90, 2018.
\bibitem{ref10} K.~G.~Liakos et al., ``Machine learning in agriculture: A review,'' \textit{Sensors}, vol.~18, no.~8, p.~2674, 2018.
\bibitem{ref11} J.~Mateo-Sagasta, S.~M.~Zadeh, and H.~Turral, \textit{More People, More Food, Worse Water?} FAO, Rome, 2018.
\bibitem{ref14} S.~P.~Mohanty, D.~P.~Hughes, and M.~Salath\'e, ``Using deep learning for image-based plant disease detection,'' \textit{Front. Plant Sci.}, vol.~7, p.~1419, 2016.
\bibitem{ref15} P.~P.~Ray, ``Internet of things for smart agriculture,'' \textit{J. Ambient Intell. Smart Environ.}, vol.~9, no.~4, pp.~395--420, 2017.
\bibitem{ref16} W.~Shi, J.~Cao, Q.~Zhang, Y.~Li, and L.~Xu, ``Edge computing: Vision and challenges,'' \textit{IEEE Internet Things J.}, vol.~3, no.~5, pp.~637--646, 2016.
\bibitem{refphys1} M.~A.~Ahmed, J.~Passioura, and A.~Carminati, ``Hydraulic processes in roots and the rhizosphere pertinent to increasing yield of water-limited grain crops: a critical review,'' \textit{J. Exp. Bot.}, vol.~69, no.~13, pp.~3255--3265, 2018.
\bibitem{refphys2} Y.~M\"ullers, J.~A.~Postma, H.~Poorter, and D.~van Dusschoten, ``Stomatal conductance tracks soil-to-leaf hydraulic conductance in faba bean and maize during soil drying,'' \textit{Plant Physiol.}, vol.~190, no.~4, pp.~2279--2294, 2022.
\end{thebibliography}
\end{document}